\documentclass[letterpaper]{article} 
\usepackage{aaai2026}  
\usepackage{times}  
\usepackage{helvet}  
\usepackage{courier}  
\usepackage[hyphens]{url}  
\usepackage{graphicx} 
\usepackage{natbib}  
\usepackage{caption} 
\usepackage{algorithm}
\usepackage{amsfonts}
\usepackage{amsmath}
\usepackage{algpseudocode}
\usepackage{booktabs}
\usepackage{newfloat}
\usepackage{listings}
\DeclareCaptionStyle{ruled}{labelfont=normalfont,labelsep=colon,strut=off} 
\floatstyle{ruled}
\newfloat{listing}{tb}{lst}{}
\floatname{listing}{Listing}
\title{DaSAThco: Data-Aware SAT Heuristics Combinations Optimization via Large Language Models}
\author{
    Minyu Chen, Guoqiang Li\thanks{Corresponding Author.}
}
\affiliations{
    School of Computer Science, Shanghai Jiao Tong University, Shanghai 200240, China\\

    \{minkow, li.g\}@sjtu.edu.cn
}

\usepackage{bibentry}

\begin{document}

\maketitle

\begin{abstract}
The performance of Conflict-Driven Clause Learning solvers hinges on internal heuristics, yet the heterogeneity of SAT problems makes a single, universally optimal configuration unattainable. While prior automated methods can find specialized configurations for specific problem families, this dataset-specific approach lacks generalizability and requires costly re-optimization for new problem types. We introduce DaSAThco, a framework that addresses this challenge by learning a generalizable mapping from instance features to tailored heuristic ensembles, enabling a train-once, adapt-broadly model. Our framework uses a Large Language Model, guided by systematically defined Problem Archetypes, to generate a diverse portfolio of specialized heuristic ensembles and subsequently learns an adaptive selection mechanism to form the final mapping. Experiments show that DaSAThco achieves superior performance and, most notably, demonstrates robust out-of-domain generalization where non-adaptive methods show limitations. Our work establishes a more scalable and practical path toward automated algorithm design for complex, configurable systems.
\end{abstract}

\section{Introduction}

The Boolean Satisfiability (SAT) problem is a cornerstone of computational complexity theory~\cite{cook2023complexity} and a problem of immense practical importance~\cite{crawford1994experimental}. Its applications are vast, with modern solvers enabling crucial advances in diverse fields such as formal verification~\cite{prasad2005survey}, planning~\cite{rintanen2012planning}, and program analysis~\cite{harris2010program}. While SAT is NP-complete, powerful solvers based on the Conflict-Driven Clause Learning (CDCL) paradigm~\cite{audemard2009predicting} can often solve massive industrial instances with remarkable efficiency. However, this success is not uniform. The performance of these solvers is critically sensitive to the internal algorithmic heuristics that guide their search through an astronomically large solution space. Developing effective heuristics has consequently been a central focus of SAT research for decades~\cite{audemard2012glucose, liang2018machine}, traditionally relying on a long and costly process of manual design and experimentation.

The challenge of designing effective heuristics has led to the development of hyper-heuristics~\cite{burke2013hyper}, which automate the process of selecting or generating algorithms. With the advent of LLMs~\cite{achiam2023gpt, comanici2025gemini}, this field has seen a surge of innovation. Frameworks like FunSearch~\cite{romera2024mathematical}, Evolution of Heuristics (EoH)~\cite{liu2024evolution}, and ReEvo~\cite{ye2024reevo} have demonstrated that LLMs can generate novel and effective heuristics for various combinatorial optimization (CO) problems by treating algorithm design as a program search task within an evolutionary framework~\cite{van2024llamea}.

However, significant research gaps remain. Firstly, prior work on CO problems like bin packing or the traveling salesman problem~\cite{liu2024llm4ad} often focuses on generating a single heuristic policy. In contrast, a CDCL solver's performance is determined by a complex combination of multiple, interacting heuristics for tasks such as restarts~\cite{audemard2012glucose, liang2018machine} and phase selection. While the recent AutoSAT framework~\cite{sun2024autosat} has made important progress by optimizing these individual components, its methodology is tailored to a specific training dataset. This process yields a single heuristic combination optimized for that particular data distribution, but the resulting configuration lacks generalizability and the entire expensive search must be repeated for each new class of problems.
Secondly, this limitation is particularly severe because the \textit{SAT problem} is not a monolith. It serves as a universal language for encoding a vast array of problems from different domains, such as Minesweeper~\cite{kaye2000minesweeper}, cryptographic analysis~\cite{soos2009extending}, and logistics planning~\cite{rintanen2012planning}. This diversity results in a problem space with enormous heterogeneity in instance characteristics, including the number of variables, clauses, and constraint density~\cite{ansotegui2009structure}. In accordance with the No-Free-Lunch theorem~\cite{wolpert2002no}, it is therefore highly improbable that any single, static combination of heuristics can achieve optimal performance across such a diverse landscape of problem instances.

To address these challenges, we propose DaSAThco (Data-Aware SAT Heuristics Combinations Optimization). Instead of seeking a single, universally optimal solver, DaSAThco's objective is to learn a generalizable mapping from instance characteristics to tailored heuristic configurations. Our framework first defines high-level \textbf{Problem Archetypes} based on statistical features of the training data. These archetypes are then used in a dual role: they guide an LLM-powered evolutionary search to generate a diverse \textbf{portfolio of specialized heuristic ensembles}, and they define the data subsets on which these ensembles are trained. Finally, we partition the instance space based on performance to create a map that enables an adaptive, data-aware selection of the best-suited ensemble for any new SAT instance.

We summarize our main contributions as follows:
\begin{itemize}
    \item We introduce DaSAThco, a novel framework that learns a portfolio of specialized solvers and an adaptive selection mechanism, shifting the paradigm from dataset-specific optimization to generalizable, data-aware algorithm design.
    \item We propose a methodology where \textbf{Problem Archetypes} are used to guide an LLM in generating a diverse portfolio of heuristic ensembles, each specialized for different instance characteristics.
    \item Through extensive experiments, we demonstrate that DaSAThco significantly outperforms baselines. Crucially, it exhibits superior \textbf{out-of-domain generalization} compared to methods that learn a single, non-adaptive configuration, validating the effectiveness of our approach.
\end{itemize}

\section{Preliminaries \& Related Work}

\subsection{SAT and CDCL-based Solvers}
Let $V = \{x_1, \dots, x_n\}$ be a finite set of Boolean variables. The corresponding set of literals is defined as $L = V \cup \{\neg v \mid v \in V\}$. A \textit{clause} $C$ is a finite subset of $L$, representing the disjunction of its literals, with the constraint that for any variable $v \in V$, both $v$ and $\neg v$ cannot be simultaneously present in $C$. A formula $F$ in Conjunctive Normal Form (CNF) is a set of clauses, $\{C_1, \dots, C_m\}$, representing their conjunction. A \textit{truth assignment} (or interpretation) is a function $\tau: V \to \{\top, \bot\}$ that maps each variable to a truth value. The satisfaction of a formula under $\tau$, denoted by $\tau \models F$, is defined hierarchically:
\begin{enumerate}
    \item A literal $l \in L$ is satisfied by $\tau$ ($\tau \models l$) if $l = v$ and $\tau(v) = \top$, or if $l = \neg v$ and $\tau(v) = \bot$.
    \item A clause $C$ is satisfied by $\tau$ ($\tau \models C$) if there exists at least one literal $l \in C$ such that $\tau \models l$.
    \item A formula $F$ is satisfied by $\tau$ ($\tau \models F$) if for all clauses $C_i \in F$, $\tau \models C_i$.
\end{enumerate}
The Boolean Satisfiability (SAT) problem is the computational task of determining whether a given CNF formula $F$ is satisfiable, i.e., whether there exists any truth assignment $\tau$ such that $\tau \models F$.

Modern SAT solvers are predominantly based on the Conflict-Driven Clause Learning (CDCL) framework. A CDCL-based solver iteratively builds a partial assignment by making decisions and applying Boolean Constraint Propagation (BCP). When a conflict arises, the solver analyzes the cause, learns a new clause to prevent the same conflict from recurring, and backtracks. This process is profoundly influenced by a complex interplay of internal heuristics. These strategies guide key aspects of the search, including which variable to decide on next (branching), when to abandon an unpromising search path and restart, and how to diversify the search by adjusting variable phases. The specific combination and implementation of these heuristics are what differentiate modern solvers and are the primary determinants of their performance.

\subsection{LLM-based Heuristic Generation}
The emergence of LLMs has opened a new frontier for automated algorithm design. By leveraging their powerful code generation and reasoning capabilities, researchers have started to automate the discovery of heuristics. A prominent approach is to frame heuristic generation as a program search problem within an evolutionary framework. FunSearch established this paradigm by evolving programs to find new mathematical discoveries~\cite{romera2024mathematical}. This was extended to CO problems by frameworks like EoH~\cite{liu2024evolution} and ReEvo~\cite{ye2024reevo}, which use evolutionary algorithms to prompt an LLM to iteratively refine and improve heuristic code for problems like TSP and online bin packing. ReEvo notably introduced a reflective step, where the LLM provides textual feedback to guide the evolutionary search, emulating a verbal gradient~\cite{ye2024reevo}. MEoH~\cite{yao2025multi} models automated heuristic design as a multi-objective optimization problem, using a dominance-dissimilarity mechanism with an LLM to generate a set of heuristics that balance performance and efficiency.

AutoSAT~\cite{sun2024autosat} was the first to apply this paradigm to the intricate environment of SAT solvers. Recognizing that generating a competitive solver from scratch is infeasible due to code complexity, AutoSAT proposed a modular framework where an LLM optimizes specific, pre-defined heuristic functions within an existing solver. It successfully demonstrated that an LLM could enhance a baseline CDCL solver to achieve competitive and, in some cases, state-of-the-art performance.

While these methods are powerful, they generally produce a single, universally applied heuristic or configuration, overlooking the instance-specific nature of algorithm performance. An emerging paradigm in automated algorithm design seeks to overcome this limitation by partitioning a problem class into subclasses based on instance features. This allows for the creation of specialized heuristics tailored to the unique characteristics of each subclass. Our work, DaSAThco, applies this principle of data-awareness to the multi-heuristic, complex environment of CDCL solvers. In doing so, we bridge the gap between the universal optimization of frameworks like AutoSAT and a more granular, data-centric approach to algorithm design.

\subsection{Algorithm Selection and Portfolio Solvers}
A major paradigm for tackling instance heterogeneity is portfolio-based algorithm selection, pioneered by the influential SatZilla framework~\cite{xu2008satzilla}. SatZilla leverages a portfolio of diverse, human-designed solvers and uses machine learning models to predict the best-performing one for a given instance based on its features. DaSAThco inherits this data-driven philosophy but introduces a fundamental novelty: rather than selecting from a portfolio of pre-existing solvers, it first uses an LLM to \textit{automatically generate} a new portfolio of fine-grained heuristic ensembles. DaSAThco is thus not only an algorithm selector but also an automated portfolio generator, a key distinction that significantly expands the space of possible solver configurations.

\section{Methodology}


\begin{figure*}[ht]
\centering
\includegraphics[width=16cm]{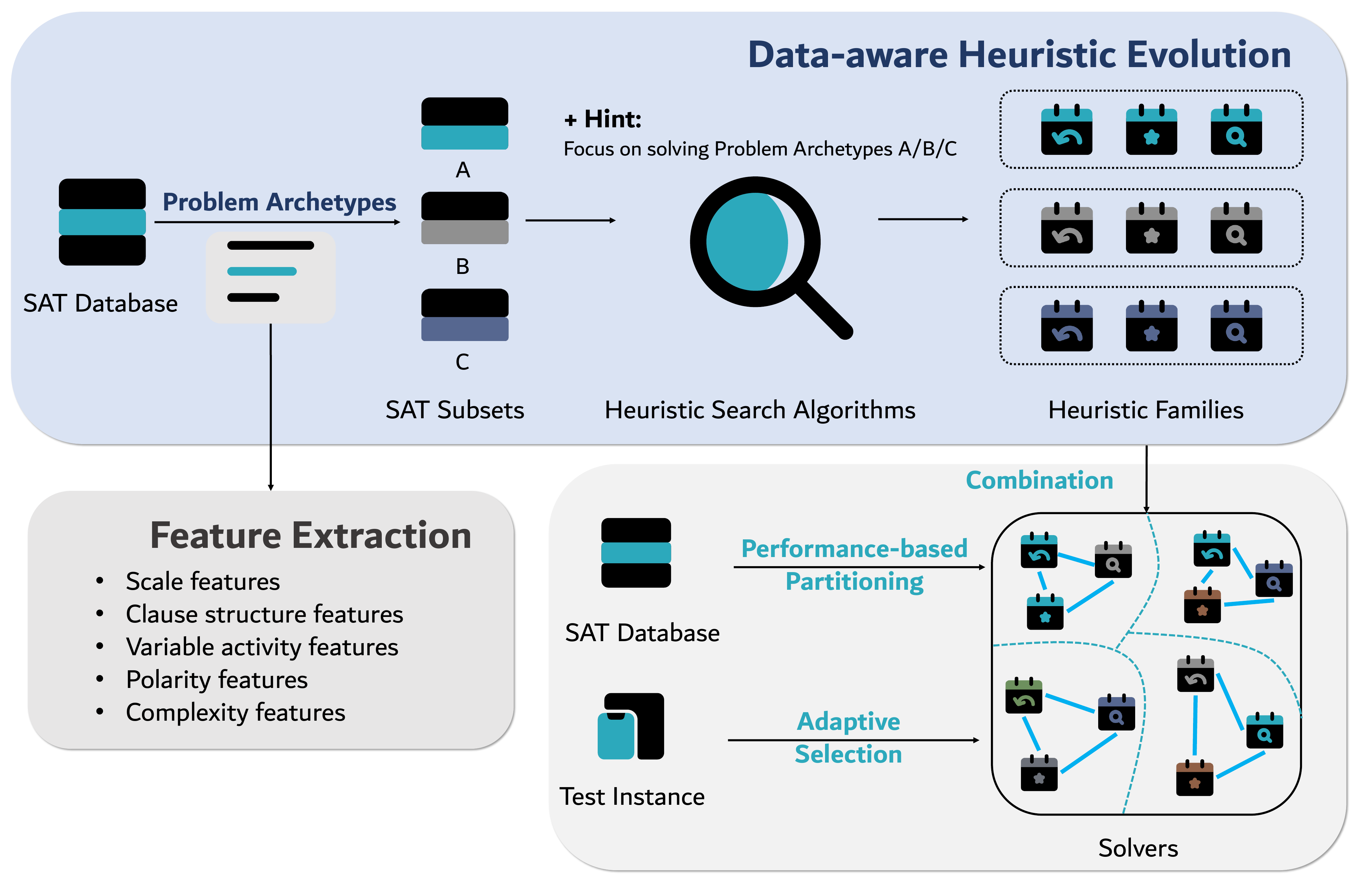}

\caption{An overview of the DaSAThco framework, illustrating the three primary stages: (1) Data-Aware Heuristic Evolution, (2) Instance Space Partitioning, and (3) Adaptive Heuristic Selection.}
\label{figx}
\end{figure*}

To address the challenge of dataset-specific optimization and the lack of generalizability in prior work, we propose DaSAThco. The core of our methodology is a paradigm shift: instead of repeatedly executing an expensive search for a single, specialized solver for each new problem family, our goal is to construct a single, robust, and adaptive framework that generalizes across them.

Specifically, DaSAThco's objective is not to find one \textit{best} heuristic ensemble, but to learn a rich \textit{mapping} from the instance feature space to the space of high-performance heuristic configurations. This is achieved by first creating a diverse portfolio of specialized heuristic ensembles and then learning an intelligent selection mechanism. Once constructed, this framework can be deployed on new, unseen SAT instances from diverse problem families, dynamically selecting a suitable configuration without the need for re-optimization. This "train-once, adapt-broadly" approach is designed to be more practical and efficient for real-world applications. The framework operates in three main stages: (1) Data-Aware Heuristic Evolution, to build the rich portfolio of heuristic components; (2) Instance Space Partitioning, to learn the mapping between instance types and optimal ensembles; and (3) Adaptive Heuristic Selection, to apply this mapping to new instances.

\subsection{Heuristic Modules in CDCL-based Solvers}
Following the modular design of AutoSAT, we focus on optimizing a set of critical, independent heuristic functions within a CDCL-based solver. For this work, we target three key heuristic categories:
\begin{enumerate}
    \item \textbf{Restart Policy (\texttt{restart}):} This module implements a crucial strategy to prevent the solver from becoming trapped in unproductive regions of the search space. A restart policy determines when to abandon the current search path and backtrack to the top decision level. While the current partial assignment is discarded, all learned clauses are retained, allowing the solver to begin a new search attempt with more information. Modern policies are often dynamic, adapting their restart frequency to the search progress to effectively combat heavy-tailed runtime distributions~\cite{luby1993optimal, audemard2012glucose}.

    \item \textbf{Phase Selection (\texttt{rephase}):} This heuristic acts as a diversification mechanism by managing the default polarity (true or false) assigned to variables. During branching, the solver often uses a variable's last assigned value as a default choice. Over time, these saved phases can lead to search stagnation. The \texttt{rephase} function is called periodically to alter these saved phases, for instance by resetting them or flipping them. This forces the solver to explore different branches first, effectively diversifying the search and pushing it into new areas of the solution space~\cite{jeroslow1990solving}.

    \item \textbf{Variable Bumping (\texttt{bump\_var}):} This module is a core component of adaptive branching strategies. When the solver encounters a conflict, it analyzes the implication graph to identify the variables responsible for the contradiction. The \texttt{bump\_var} function is then invoked to increase a numerical score, often called an `activity score', associated with each of these variables. By elevating the scores of conflict-prone variables, this mechanism dynamically influences the main branching heuristic to prioritize them in future decisions. This effectively focuses the search on the most constrained and active parts of the problem space~\cite{biere2009conflict, liang2016learning}.

\end{enumerate}
Our goal is to find not just one optimal implementation for each of these, but a diverse set of effective implementations that can be combined into powerful ensembles.


\subsection{Feature Extraction for SAT Instances}
A critical precursor to any data-aware method is the ability to characterize instances with a quantitative feature vector. We define a 21-dimensional feature vector $v(j)$ that maps a given SAT instance $j$ to a vector in $\mathbb{R}^{21}$, designed to capture a comprehensive set of its structural and statistical properties. The features are organized into five distinct categories: (1) basic scale features, such as the number of variables and clauses; (2) clause structure features describing the length and type of clauses (e.g., unit, binary); (3) variable activity features that measure the distribution and frequency of variables; (4) polarity features to capture the balance of positive and negative literals; and (5) overall complexity metrics like constraint density. This comprehensive signature enables robust downstream clustering and similarity comparisons. A detailed breakdown of all 21 features is provided in the Appendix.

\subsection{Data-Aware Heuristic evolution}
To build our library of heuristics, we introduce a guided search process centered around a set of predefined \textit{optimization directions}. This approach makes the data-aware evolution process more explicit and controllable. The total set of possible heuristic ensembles is the Cartesian product $\mathcal{H} = \mathcal{L}_{\text{restart}} \times \mathcal{L}_{\text{rephase}} \times \mathcal{L}_{\text{bump\_var}}$. Before proceeding to the main partitioning stage, this large set of candidate ensembles undergoes a pruning step, where combinations that exhibit poor performance on a general benchmark subset are filtered out, resulting in a smaller, high-quality portfolio $\mathcal{H'}$.



\textbf{Defining Problem Archetypes and Search Environments.}
We first establish a set of high-level, interpretable \textbf{Problem Archetypes} that represent distinct categories of SAT instances. This allows our framework to move beyond generic optimization and target specific problem structures. For example, an archetype can be defined for instances with an \textbf{extremely high clause-to-variable ratio}, representing highly-constrained problems where the search for satisfying assignments is exceptionally tight. Another archetype could target \textbf{large-scale problems} with both high numbers of clauses and variables, which poses a significant challenge to the scalability of any search strategy. A third, more nuanced archetype might focus on \textbf{problems with a heterogeneous clause structure}, characterized by a large standard deviation in clause length and a mix of many short and very long clauses, a common feature in industrial benchmarks. While these archetypes are conceptual, they are systematically defined by discretizing key metrics from our feature vector.

These archetypes serve a crucial dual role in creating specialized search environments. For each archetype $d_{i}$, we curate a corresponding data subset $I_{i}\subseteq I_{train}$ by filtering instances that match its description. This specialized subset becomes the dedicated training and evaluation ground for evolving a heuristic with a specific performance preference, while the textual archetype $d_{i}$ simultaneously guides the LLM's creative process toward that same preference.

\textbf{Guided Evolutionary Search.}
Let $\mathcal{D} = \{(d_i, I_i)\}_{i=1}^p$ be the set of established pairs, where each $d_i$ is an optimization direction and $I_i$ is its corresponding data subset. For each pair $(d_i, I_i) \in \mathcal{D}$, we conduct an independent evolutionary search for each heuristic module. During this search, the textual direction $d_i$ is injected as a guiding hint into the LLM's prompt. This guidance mechanism is a core contribution of our work; it is not a new search algorithm, but rather a modular and compatible \textbf{plug-in} working at data-level designed to enhance existing LLM-based evolutionary frameworks such as EoH and EPS~\cite{zhang2024understanding}. The prompt thus directs the LLM to generate a heuristic specialized for direction $d_i$, with its performance evaluated exclusively on the dedicated data subset $I_i$.

\subsection{Instance Space Partitioning via Performance-Based Clustering}
The Cartesian product of generated heuristics can yield a large number of candidate ensembles, posing a computational challenge for the partitioning stage. To manage this, we constrain the number of components per heuristic type (e.g., $k \le 3$) and perform an aggressive pruning step. All generated ensembles are evaluated on a benchmark subset, and only a fixed number of the top-performing candidates are retained in the final portfolio, $\mathcal{H'}$.

Given this pruned portfolio of high-quality ensembles $\mathcal{H'}$, the next stage is to understand which ensemble is best for which kind of instance. To achieve this, we partition our training set of instances, $I_{train}$, based on performance.

First, we evaluate every heuristic ensemble $h_i \in \mathcal{H}$ on every instance $j \in I_{train}$. Let $p(h_i, j)$ be the performance metric (e.g., PAR-2 score) of ensemble $h_i$ on instance $j$.

Next, for each instance $j$, we identify its optimal ensemble, $h^*(j)$, from our library:
$$h^*(j) = \arg\min_{h_i \in \mathcal{H}} p(h_i, j)$$
This allows us to partition the instance set $I_{train}$ into disjoint clusters, where each cluster $C_i$ is associated with a single best-performing ensemble $h_i$:
$$C_i = \{j \in I_{train} \mid h^*(j) = h_i\}$$
This process creates a direct mapping from a region in the instance feature space (represented by the instances in $C_i$) to an optimal solver configuration $h_i$. For each resulting cluster $C_i$, we compute its feature space centroid, $\bar{v}_i$, by averaging the feature vectors of all its member instances.

\subsection{Adaptive Heuristic Selection for Instances}
With the partitioned instance space and associated optimal ensembles, we can now perform adaptive selection for any new, unseen test instance $j_{new}$. This process is summarized in Algorithm \ref{alg:dasathco_framework}.

When a new instance $j_{new}$ arrives, we first extract its feature vector $v(j_{new})$. We then calculate the distance (using normalized Euclidean distance) between $v(j_{new})$ and each cluster centroid $\bar{v}_i$. The heuristic ensemble $h_k$ associated with the closest centroid $\bar{v}_k$ is selected as the most suitable configuration for solving $j_{new}$. This allows the solver to dynamically adapt its strategy based on the data characteristics of the problem at hand.

\begin{algorithm}[H] 
\caption{DaSAThco Framework Overview}
\label{alg:dasathco_framework}
\begin{algorithmic}[1]
\State \textbf{Input:} Training instance set $I_{train}$, test instance $j_{new}$.
\State \textbf{Output:} Solved result for $j_{new}$.

\Comment{Stage 1: Data-Aware Heuristic Evolution}
\State Define a set of Problem Archetypes $D=\{d_{1},...,d_{p}\}$.
\For{each archetype $d_{i}\in D$}
    \State Create data subset $I_{i}\subseteq I_{train}$ by filtering.
\EndFor
\State $\mathcal{H}\leftarrow\mathcal{L}_{restart}\times\mathcal{L}_{rephase}\times\mathcal{L}_{bump\_var}$
\State $\mathcal{H'} \leftarrow \text{Prune}(\mathcal{H}, I_{train})$ by removing ensembles with performance below a predefined threshold.

\Comment{Stage 2: Offline Instance Space Partitioning}
\For{each instance $j\in I_{train}$}
    \State $h^{*}(j)\leftarrow \arg\min_{h_{k}\in \mathcal{H'}} \text{performance}(h_{k},j)$.
\EndFor
\For{each unique $h_{k}$ that is optimal for some instance}
    \State $C_{k}\leftarrow\{j\in I_{train}|h^{*}(j)=h_{k}\}$.
    \State $\overline{v}_{k}\leftarrow\frac{1}{|C_{k}|}\sum_{j\in C_{k}}v(j)$.
\EndFor

\Comment{Stage 3: Online Adaptive Selection}
\State $v_{new}\leftarrow v(j_{new})$.
\State $k \leftarrow \arg\min_i \text{distance}(v_{new}, \bar{v}_i)$.
\State $h_{selected}\leftarrow h_{k}$.
\State Solve $j_{new}$ using the solver configured with $h_{selected}$.
\end{algorithmic}
\end{algorithm}

\section{Experiments}
To evaluate the effectiveness of DaSAThco, we conduct a series of experiments designed to assess its performance against baseline and state-of-the-art SAT solvers.

\subsection{Experimental Setup}
\textbf{Environment and Parameters.}
All solvers are implemented in C++ and compiled with g++ 12.3.0. The LLM interaction and the evolutionary framework are managed in Python. Experiments were conducted on servers equipped with AMD Ryzen 9 5950X 16-core processors and 128GB of RAM. For all heuristic generation tasks, we utilized the GPT-4o model with a temperature of 0.8 to encourage diverse outputs. Considering the computational expense of solving SAT problems and the scale of our benchmarks, the timeout for solving any single SAT instance was set to 1000 seconds.

\textbf{Backbone Solver.}
To ensure a methodologically consistent and comparable basis, we follow the precedent set by AutoSAT~\cite{sun2024autosat} in selecting our backbone solver. Our framework is built upon \textit{EasySAT}, a lightweight and modular CDCL solver. As established in prior work, this choice provides a clean and capable baseline that is well-suited for modification by LLMs, striking a practical balance between solver functionality and the token-context limitations inherent in current LLM-based code generation.

\textbf{Datasets.}
Our experimental design aims to evaluate the generalization capability of DaSAThco. We construct a single, heterogeneous \textbf{training set} composed of instances from multiple sources, including unclassified problems from the SAT Competitions of 2022 and 2023. Our evaluation is then divided into two settings:
\begin{itemize}
    \item \textbf{In-Domain Generalization:} We test on unseen instances from families that were partially represented in the training set. This includes instances from the \textit{CoinsGrid}, \textit{LangFord}, and \textit{PRP (Profitable Robust Production)} benchmarks.
    \item \textbf{Out-of-Domain Generalization:} To evaluate true generalization capabilities, we use entire problem families that were completely held out during the heuristic evolution process. These OOD test sets include the \textit{CNP (Chromatic Number of the Plane)}, \textit{Zamkeller}, and \textit{Knightour} datasets.
\end{itemize}

\textbf{Baselines.}
We compare the performance of DaSAThco against a set of representative baselines:
\begin{itemize}
    \item \textbf{EasySAT}: The lightweight, modular CDCL solver that serves as the direct backbone for our modifications. Its performance represents the starting point before any LLM-based optimization.
    \item \textbf{AutoSAT}~\cite{sun2024autosat}: A state-of-the-art framework representing the prior paradigm of dataset-specific optimization. To ensure a fair comparison of generalization capabilities, we adapt its methodology to our experimental setting. Instead of running its search process on each individual problem family, we run AutoSAT on the same single, heterogeneous training set used by DaSAThco. This evaluates its ability to find a single "average-best" configuration for a diverse set of problems. Consequently, the performance reported here is not directly comparable to its original publication, where it was optimized on specialized datasets, and may be lower.
    \item \textbf{MiniSat}: A classic and highly-optimized CDCL solver, which serves as a robust and widely-recognized traditional baseline.
\end{itemize}

\begin{table*}[t]
\centering
\caption{Main performance comparison across In-Domain and Out-of-Domain datasets. For each solver, we report both the PAR-2 score (lower is better) and the number of solved instances (Solved, higher is better). The timeout is 1000s. Best results in each category are in bold.}
\label{tab:main_results}
\begin{tabular}{l@{\hspace{1.5em}}r@{\hspace{1.5em}}rr@{\hspace{2em}}rr@{\hspace{2em}}rr@{\hspace{2em}}rr}
\toprule
& & \multicolumn{2}{c}{\textbf{EasySAT}} & \multicolumn{2}{c}{\textbf{MiniSat}} & \multicolumn{2}{c}{\textbf{AutoSAT}} & \multicolumn{2}{c}{\textbf{DaSAThco}} \\
\midrule
\textbf{Dataset} & \textbf{\# Inst.} & PAR-2 & Solved & PAR-2 & Solved & PAR-2 & Solved & PAR-2 & Solved \\
\midrule
\multicolumn{10}{c}{\textit{In-Domain Benchmarks}} \\
\midrule
CoinsGrid & 52 & 1757.5 & 7 & 1895.1 & 4 & 1711.1 & 9 & \textbf{1671.0} & \textbf{10} \\
LangFord  & 64 & 1915.9 & 4 & 1995.4 & 0 & 1850.2 & 8 & \textbf{1750.9} & \textbf{12} \\
PRP       & 144& 1970.4 & 3 & 1935.8 & 9 & 1843.8 & 15 & \textbf{1739.6} & \textbf{20} \\
\midrule
\multicolumn{10}{c}{\textit{Out-of-Domain Benchmarks}} \\
\midrule
CNP       & 50 & 594.9 & 38 & 1150.2 & 22 & 614.4 & 39 & \textbf{550.7} & \textbf{41} \\
Zamkeller & 48 & 830.5 & 30 & 1650.6 & 5 & 764.8 & 32 & \textbf{611.4} & \textbf{35} \\
KnightTour& 22 & 1733.4 & 3 & 1800.7 & 2 & 1684.6 & 4 & \textbf{1615.6} & \textbf{5} \\
\bottomrule
\end{tabular}
\end{table*}

\textbf{Metrics.}
We evaluate solver performance using two standard metrics in the SAT competition: the number of solved instances within the timeout, and the Penalized Average Runtime with a factor of 2 (PAR-2) score. The PAR-2 score is the average runtime across a set of instances, but with a heavy penalty for any instance that is not solved within the 1000s timeout. Specifically, unsolved instances are assigned a runtime of twice the timeout (2000s). The score is calculated as:
$$ \text{PAR-2} = \frac{1}{N} \sum_{i=1}^{N} t_i' $$
where $N$ is the number of instances, and $t_i'$ for instance $i$ is its actual runtime if solved, or 2000s if unsolved. A lower PAR-2 score is better, as it indicates a solver is both fast and robust. Crucially, the PAR-2 score also serves as the primary fitness metric that guides the LLM-driven evolutionary search for better heuristics.


\subsection{Performance Comparison}
We evaluate the performance of DaSAThco against the baselines on both in-domain and out-of-domain datasets. The detailed results are presented in Table~\ref{tab:main_results}. The findings clearly demonstrate the superiority of our data-aware, portfolio-based approach.

Overall, DaSAThco consistently outperforms all baselines across both evaluation settings. On the in-domain datasets (CoinsGrid, LangFord, and PRP), DaSAThco achieves the best PAR-2 scores and solves the most instances. The improvement is particularly notable on the complex PRP dataset, where DaSAThco solves 20 instances, a significant increase compared to the 15 solved by AutoSAT. This indicates that even for familiar problem archetypes, DaSAThco's adaptive selection from a specialized portfolio is more effective than relying on a single, average-best configuration.

The strength of DaSAThco's generalization capability is most evident on the OOD datasets, which represent problem families completely unseen during the heuristic evolution phase. On all three OOD datasets (CNP, Zamkeller, and KnightTour), DaSAThco delivers the best performance. For instance, on Zamkeller, it not only achieves the lowest PAR-2 score but also solves more instances than both AutoSAT and the traditional baselines. This robust performance on novel problem structures stands in contrast to the more brittle generalization of the non-adaptive methods and strongly supports our central thesis: by learning a mapping between instance features and a portfolio of specialized heuristics, DaSAThco can adapt to new challenges more effectively than methods that converge on a single configuration.

\subsection{Ablation Study}
To understand the contribution of the key components of our framework, we conduct an ablation study on two representative datasets: \textit{CoinsGrid} for in-domain generalization, and \textit{Zamkeller} for out-of-domain (OOD) generalization. We compare our full DaSAThco model against three ablated variants:
\begin{itemize}
    \item \textbf{w/o Data-Aware Generation:} The portfolio is generated without the guidance of Problem Archetypes or the use of specialized data subsets. The adaptive selection mechanism is then applied to this non-specialized portfolio.
    \item \textbf{Random Selection:} For each test instance, we randomly select a heuristic ensemble from the fully generated, specialized portfolio.
    \item \textbf{Single Best Selection:} We select the single heuristic ensemble that performs best on average across the entire training set and apply it to all test instances, simulating the non-adaptive paradigm of prior work.
\end{itemize}

The results are presented in Table~\ref{tab:ablation_results}. The full DaSAThco model achieves the best performance on both datasets. The contribution of our data-aware generation is evident when comparing the full model to the variant without the data-aware generation process, as the latter shows a clear performance drop.

The analysis of the selection mechanism reveals a crucial insight. On the OOD dataset \textit{Zamkeller}, the single best selection approach performs significantly worse than the adaptive DaSAThco, highlighting the failure of a non-adaptive strategy to generalize to unseen problem archetypes. Interestingly, on the in-domain \textit{CoinsGrid} dataset, the single best variant proves to be a strong baseline. This suggests that for familiar problem types, a well-tuned single solver can be effective. Nevertheless, DaSAThco's adaptive selection still yields the best results on both datasets, demonstrating its superior ability to pick the optimal configuration. Both the data-aware portfolio generation and the dynamic selection mechanism are therefore critical to DaSAThco's robust performance.

\begin{table}[h]
\centering
\caption{Ablation study on a representative In-Domain (CoinsGrid) and Out-of-Domain (Zamkeller) dataset. We report the PAR-2 score (lower is better) and the number of solved instances (\#S, higher is better).}
\label{tab:ablation_results}
\begin{tabular}{l@{\hspace{1.5em}}rr@{\hspace{2em}}rr}
\toprule
& \multicolumn{2}{c}{\textbf{CoinsGrid}} & \multicolumn{2}{c}{\textbf{Zamkeller}} \\
\midrule
\textbf{Model Variant} & PAR-2 & \#S & PAR-2 & \#S \\
\midrule
\textbf{DaSAThco} & \textbf{1671.0} & \textbf{10} & \textbf{611.4} & \textbf{35} \\
w/o Data-Aware & 1735.8 & 9 & 722.4 & 33 \\
Random & 1740.2 & 8 & 780.5 & 29 \\
Single Best & 1711.1 & 9 & 764.8 & 32 \\
\bottomrule
\end{tabular}
\end{table}

\subsection{Analysis of Portfolio Scale and Diversity}
The size of the heuristic portfolio is a critical hyperparameter in our framework. To analyze its impact while maintaining a focused experimental scope, we first conducted preliminary investigations which suggested that the \texttt{rephase} heuristic had a comparatively smaller impact on overall performance across different problem archetypes. Therefore, we fixed the \texttt{rephase} component to its default implementation and focused our sensitivity analysis on the two more influential modules: \texttt{restart} and \texttt{bump\_var}. We evaluated the performance of DaSAThco on the \textit{CoinsGrid} dataset while varying the portfolio cardinalities for these two heuristics ($k_{restart}$ and $k_{bump\_var}$), with k ranging from 0 to 3.

\begin{figure}[h]
    \centering
    \includegraphics[width=0.95\columnwidth]{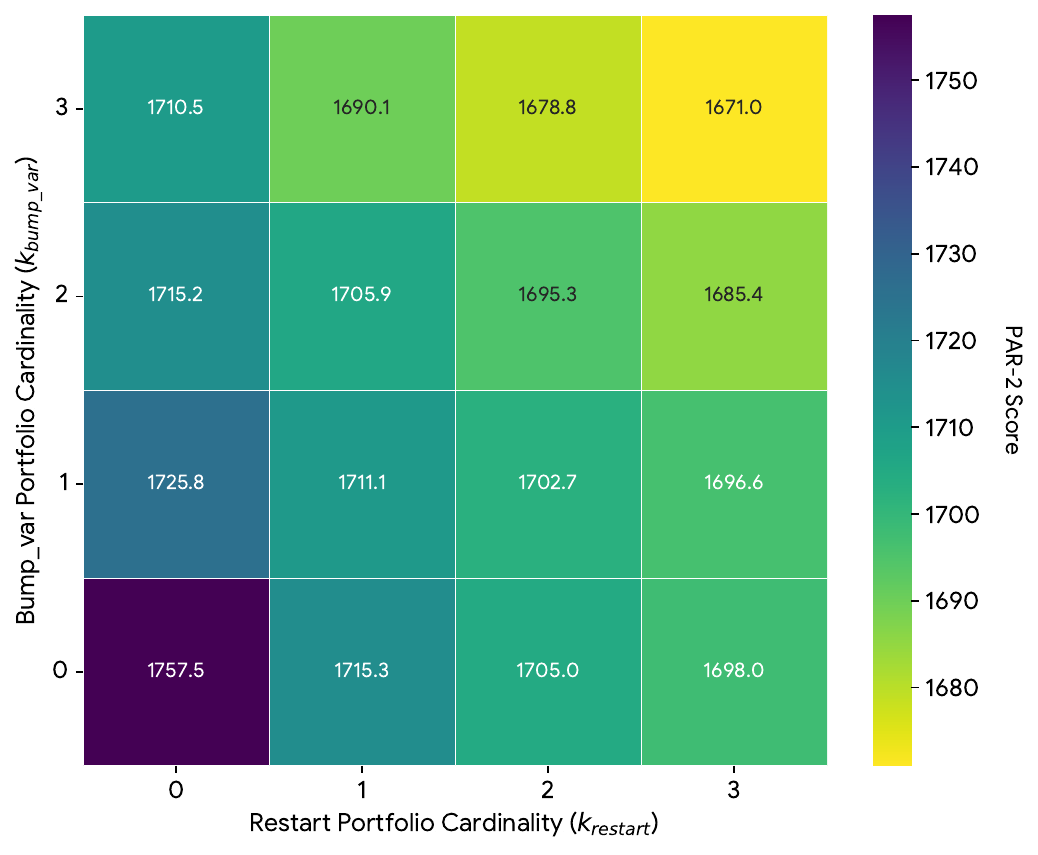}
    \caption{Sensitivity analysis of DaSAThco's performance on the CoinsGrid dataset. The heatmap shows the average PAR-2 score (lower is better) as a function of the portfolio cardinalities for the \texttt{restart} ($k_{restart}$) and \texttt{bump\_var} ($k_{bump\_var}$) heuristics. The k=0 point corresponds to the EasySAT baseline.}
    \label{fig:sensitivity_heatmap}
\end{figure}

Figure~\ref{fig:sensitivity_heatmap} presents a heatmap of the results, which reveals two key insights. First, the most significant performance gain occurs when moving from k=0 (the EasySAT baseline) to k=1. This sharp improvement demonstrates that the heuristic functions discovered by the LLM can effectively enhance the original solver's efficiency, thus validating the fundamental feasibility of our LLM-based search approach. Second, as the portfolio cardinalities increase from 1 to 3, the performance continues to improve, but the changes are more gradual. This illustrates a boundary effect, where the marginal benefit of adding more heuristic diversity diminishes. Nevertheless, the optimal performance is achieved with the largest tested portfolio ($k_{restart}=3, k_{bump\_var}=3$), confirming that a richer, more diverse portfolio is beneficial within the tested range.


\subsection{Overhead Analysis}
We analyze the computational overhead of DaSAThco in terms of its one-time offline costs and per-instance online costs. The offline cost involves generating and compiling the heuristic portfolio, a process made manageable by the lightweight EasySAT backbone. The per-instance online overhead for solving a new instance is negligible, consisting of two fast operations: feature extraction and adaptive selection. The calculation of the 21-dimensional feature vector is computationally inexpensive and can be pre-computed for known benchmarks. Subsequently, our adaptive selection via a nearest-centroid search is extremely efficient, consistently taking less than a second. This minimal online cost makes DaSAThco highly practical. Future work could explore the trade-off between this efficient selector and more sophisticated but potentially slower models, such as using an LLM for the selection task itself.

\section{Conclusion}
In this work, we introduced DaSAThco, a novel framework that addresses the critical generalizability limitations in automated SAT solver design by shifting from dataset-specific optimization to a scalable "train-once, adapt-broadly" paradigm. Our methodology leverages a Large Language Model, guided by Problem Archetypes, to generate a diverse portfolio of specialized heuristic ensembles and then learns an adaptive mechanism to select the best configuration for new instances. Experiments confirm that this approach not only improves performance but, more importantly, exhibits robust out-of-domain generalization. This validates that learning a mapping from instance features to a generated portfolio of solvers is a more effective and practical paradigm for the automated design of complex, configurable systems like SAT solvers. Future work could focus on automating the discovery of these archetypes and extending this paradigm to other domains.

\bibliography{aaai2026}

\appendix
\section{Detailed SAT Instance Features}
\label{app:features}
The 21 features used to construct the feature vector for each SAT instance are detailed below, organized by category.

\begin{itemize}
    \item \textbf{Basic Features:} These capture the overall size and scale of the instance.
    \begin{itemize}
        \item \texttt{num\_variables}: The number of variables ($|V|$).
        \item \texttt{num\_clauses}: The number of clauses ($|C|$).
        \item \texttt{var\_clause\_ratio}: The ratio of variables to clauses ($|V|/|C|$).
    \end{itemize}
    \item \textbf{Clause Structure Features:} These describe the properties of the clauses in the formula.
    \begin{itemize}
        \item \texttt{avg\_clause\_length}, \texttt{std\_clause\_length}, \texttt{min\_clause\_length}, \texttt{max\_clause\_length}: Statistical measures of the number of literals per clause.
        \item \texttt{clause\_length\_entropy}: The entropy of the clause length distribution, measuring its diversity.
        \item \texttt{unit\_clause\_ratio}, \texttt{binary\_clause\_ratio}, \texttt{long\_clause\_ratio}: The proportion of clauses that are unit (length 1), binary (length 2), or long (length $\ge$ 5).
    \end{itemize}
    \item \textbf{Variable Activity Features:} These describe how variables are distributed across clauses.
    \begin{itemize}
        \item \texttt{avg\_var\_frequency}, \texttt{std\_var\_frequency}, \texttt{max\_var\_frequency}: Statistical measures of how many times each variable appears in the formula.
        \item \texttt{var\_frequency\_entropy}: The entropy of the variable frequency distribution.
        \item \texttt{singleton\_var\_ratio}: The ratio of variables that appear in only one clause.
    \end{itemize}
    \item \textbf{Polarity Features:} These capture the balance between positive and negative literals.
    \begin{itemize}
        \item \texttt{positive\_literal\_ratio}: The overall ratio of positive literals to total literals.
        \item \texttt{balanced\_var\_ratio}: The proportion of variables that appear an equal number of times positively and negatively.
        \item \texttt{pure\_literal\_ratio}: The proportion of variables that appear with only one polarity.
        \item \texttt{polarity\_bias}: A measure of the overall tendency towards positive or negative literals.
    \end{itemize}
    \item \textbf{Complexity Features:}
    \begin{itemize}
        \item \texttt{constraint\_density}: A measure of how constrained the problem is, defined as the total number of literals divided by the product of variables and clauses.
    \end{itemize}
\end{itemize}

\section{Dataset Characteristics}
\label{app:datasets}
Table~\ref{tab:dataset_stats} provides a statistical overview of the datasets used in our evaluation. The instance counts correspond to the specific subsets used in our experiments, while the variable and clause statistics are sourced from prior work~\cite{sun2024autosat} to provide context on their general characteristics.

\begin{table*}[!ht]
\centering
\caption{Statistical characteristics of the evaluation datasets.}
\label{tab:dataset_stats}
\begin{tabular}{lrrr}
\toprule
\textbf{Dataset} & \textbf{\# Inst.} & \textbf{Variables (Mean $\pm$ Std)} & \textbf{Clauses (Mean $\pm$ Std)} \\
\midrule
\multicolumn{4}{c}{\textit{In-Domain Benchmarks}} \\
\midrule
CoinsGrid & 52 & 530807 $\pm$ 513663 & 3825868 $\pm$ 3701594 \\
LangFord  & 64 & 312492 $\pm$ 213284 & 2734786 $\pm$ 1972624 \\
PRP       & 144 & 499206 $\pm$ 324889 & 3337426 $\pm$ 2175367 \\
\midrule
\multicolumn{4}{c}{\textit{Out-of-Domain Benchmarks}} \\
\midrule
CNP       & 50 & 9890 $\pm$ 11139 & 86724 $\pm$ 77379 \\
Zamkeller & 48 & 21435 $\pm$ 19119 & 265218 $\pm$ 283330 \\
KnightTour& 22 & 135288 $\pm$ 191062 & 5742107 $\pm$ 9872215 \\
\bottomrule
\end{tabular}
\end{table*}

\section{Hyperparameter Settings}
\label{app:hyperparams}
Table~\ref{tab:hyperparams} lists the main hyperparameters used across our experiments to ensure reproducibility.

\begin{table}[!ht]
\centering
\caption{Main hyperparameters used in our experiments.}
\label{tab:hyperparams}
\begin{tabular}{ll}
\toprule
\textbf{Hyperparameter} & \textbf{Value} \\
\midrule
\multicolumn{2}{l}{\textit{General Experimental Settings}} \\
Solver Timeout & 1000s \\
Random Seed & 42 \\
\midrule
\multicolumn{2}{l}{\textit{LLM and Evolutionary Search}} \\
LLM Model & GPT-4o \\
Temperature & 0.8 \\
Generations & 3 \\
Population Size & 2 \\
\midrule
\multicolumn{2}{l}{\textit{DaSAThco Framework Settings}} \\
Number of Problem Archetypes & 3 \\
Heuristics per Type ($k$) & 3 \\
Pruned Portfolio Size & 6 \\
\bottomrule
\end{tabular}
\end{table}\textbf{}

\end{document}